\documentclass{article}

     \usepackage[preprint]{neurips_2022}

\usepackage[utf8]{inputenc} % allow utf-8 input
\usepackage[T1]{fontenc}    % use 8-bit T1 fonts
\usepackage{hyperref}       % hyperlinks
\usepackage{url}            % simple URL typesetting
\usepackage{booktabs}       % professional-quality tables
\usepackage{amsfonts}       % blackboard math symbols
\usepackage{nicefrac}       % compact symbols for 1/2, etc.
\usepackage{microtype}      % microtypography
\usepackage{xcolor}         % colors

\usepackage{amsmath,amsfonts,amsthm}
\usepackage{bbm}
\usepackage{amssymb}
\usepackage{algorithm}
\usepackage{algpseudocode}
\usepackage{graphicx}
\usepackage{wrapfig}

\title{On the Normalizing Constant of  \\ the Continuous Categorical Distribution}

\author{%
  Elliott Gordon-Rodriguez\thanks{Equal contribution.} \\
  Columbia University\\
  \texttt{eg2912@columbia.edu} \\
   \And
   Gabriel Loaiza-Ganem\footnotemark[1] \\
   Layer6 AI \\
   \texttt{gabriel@layer6.ai} \\
   \AND
   Andres Potapczynski \\
   New York University \\
   \texttt{andpotap@nyu.edu} \\
   \And
   John P. Cunningham \\
   Columbia University \\
   \texttt{jpc2181@columbia.edu} \\
}

\begin{document}

\maketitle

\begin{abstract}
Probability distributions supported on the simplex enjoy a wide range of applications across statistics and machine learning.
Recently, a novel family of such distributions has been discovered: the \emph{continuous categorical}.
This family enjoys remarkable mathematical simplicity; its density function resembles that of the Dirichlet distribution, but with a normalizing constant that can be written in closed form using elementary functions only.
In spite of this mathematical simplicity, our understanding of the normalizing constant remains far from complete.
In this work, we characterize the numerical behavior of the normalizing constant and we present theoretical and methodological advances that can, in turn, help to enable broader applications of the continuous categorical distribution.
Our code is available at \url{https://github.com/cunningham-lab/cb_and_cc/}.
\end{abstract}

\section{Introduction} \label{sec:intro}

The \emph{continuous categorical} (CC) distribution is defined by the following density function \citep{gordon2020continuous}:
  \begin{align} \label{eq:cc}
\bold x \sim \mathcal{CC}(\boldsymbol \lambda) \iff 
p( \bold x ; \boldsymbol \lambda) 
%= \frac{1}{C(\boldsymbol \lambda)} 
\propto \prod_{i=1}^K \lambda_i^{x_i}.
\end{align}
Here, $\bold x $ denotes a simplex-valued random variable, and $\boldsymbol \lambda$ denotes a simplex-valued parameter, in other words:\footnote{Note that the $K$-simplex is also commonly defined as $\Delta^K = \{\bold x \in \mathbb R^K_+ : \sum_{i=1}^K = 1 \}$. The two definitions are equivalent, however $\mathbb S^K$ is a subset of $\mathbb R^{K-1}$ with positive Lebesgue measure, whereas $\Delta^K$ is a subset of $\mathbb R^K$ with zero Lebesgue measure. For this reason, using $\mathbb S^K$ will facilitate our later arguments involving integrals on the simplex.}
\begin{align}
\bold x, \boldsymbol \lambda \in \mathbb S^K := \left\{ \boldsymbol x \in \mathbb R_+^{K-1} : \sum_{i=1}^{K-1} x_i \le 1 \right\},
\end{align}
where we additionally define the $K$th coordinates as the remainder:
\begin{align}
x_K &= 1 - \sum_{i=1}^{K-1} x_i %x_1 - x_2 - \dots - x_{K-1}
\\
\lambda_K &= 1 - \sum_{i=1}^{K-1} \lambda_i. %\eta_1 - \eta_2 - \dots - \eta_{K-1}.
\end{align}
It is natural to contrast the CC with the similar-looking Dirichlet distribution:
  \begin{align} \label{eq:dirichlet}
\bold x \sim \mathrm{Dirichlet}(\boldsymbol \alpha) \iff 
p( \bold x ; \boldsymbol \alpha ) \propto %\frac{\Gamma(\sum_{i=1}^K \alpha_i)}{\prod_{i=1}^K\Gamma(\alpha_i)} 
%\frac{1}{B(\boldsymbol \alpha) } 
\prod_{i=1}^K x_i^{\alpha_i - 1},
\end{align}
where again $\bold x \in \mathbb S^K$, but $\boldsymbol \alpha \in \mathbb R_+^K$ now denotes a positive unconstrained parameter vector with one more dimension than $\bold x$.

While the densities in Eq. \ref{eq:cc} and Eq. \ref{eq:dirichlet} look similar, they hide very different normalizing constants. In the Dirichlet case, it is well known that \citep{dirichlet1839nouvelle}:
\begin{align} \label{eq:dirnorm}
\int_{\mathbb S^K} \prod_{i=1}^K x_i^{\alpha_i - 1} d\mu (\bold x) = \frac{\prod_{i=1}^K \Gamma (\alpha_i) }{\Gamma(\sum_{i=1}^K \alpha_i)},
\end{align}
where $\Gamma(\alpha) = \int_0^\infty t^{\alpha - 1} e^{-t} dt$ denotes the gamma function and $\mu(\cdot)$ is the Lebesgue measure. On the other hand, the normalizing constant of the CC admits the following closed form \citep{gordon2020continuous}:
\begin{align} \label{eq:intCC}
\int_{\mathbb S^K} \prod_{i=1}^K \lambda_i^{x_i} d\mu (\bold x) =
\sum_{k=1}^K \frac
 { \lambda_k }
 {  \prod_{i\ne k} \log{\frac{\lambda_{k}}{\lambda_{i}}} },
\end{align}
which contains elementary operations only.
In spite of its mathematical simplicity, this normalizing constant can be numerically hard to compute, particularly in high dimensions.
Moreover, Eq. \ref{eq:intCC} breaks down under equality of parameters, i.e., whenever $\lambda_i = \lambda_k$ for some $i \ne k$, because the denominator evaluates to zero.
These issues will be the primary focus of our exposition, in particular:
\begin{itemize}
 \item In Section \ref{sec:catastrophic}, we characterize the numerical behavior of our normalizing constant. We demonstrate that vectorized computation can suffer from catastrophic cancellation, the severity of which depends on the proximity between parameter values. 
 \item In Section \ref{sec:laplace}, we rederive the normalizing constant as an inverse Laplace transform, which in turn can be be evaluated using numerical inversion algorithms. We show that this alternative computation strategy exhibits good numerical behavior in the regime where catastrophic cancellation is most severe.
 \item In Section \ref{sec:inductive}, we propose an orthogonal computational approach based on a recursive property of the normalizing constant.
 \item In Section \ref{sec:general}, we generalize Eq. \ref{eq:intCC} to arbitrary parameter values, i.e., including equality of parameters $\lambda_i = \lambda_k$ for any $i\ne k$. The resulting formula depends on an expectation that can be computed using automatic differentiation.
\end{itemize}

We conclude this section with some remarks.
First, note that in the 1-dimensional case, the CC distribution reduces to the \emph{continuous Bernoulli} distribution \citep{loaiza2019continuous}, which arose in the context of generative models of images \citep{kingma2013auto, bond2021deep} and provided the original inspiration for the CC family.
More generally, the CC is closely related to the categorical cross-entropy loss commonly used in machine learning \citep{gordon2020uses}.
%While this is not our focus, we note that this distribution has applications in compositional data analysis and machine learning \cite{CC, ICNBIN}.
%In the 1-dimensional case,A notable special case of the CC is the continuous Bernoulli \cite{loaiza2019continuous}, which corresponds to $K=1$. arises for modeling natural images \cite{VAE}.

We also note that the CC can be rewritten using the exponential family canonical form:
  \begin{align} \label{eq:canon}
\bold x \sim \mathcal{CC}(\boldsymbol \eta) \iff 
p( \bold x ; \boldsymbol \eta) 
= \frac{1}{C(\boldsymbol \eta)} 
 e^{\boldsymbol \eta ^ \top \bold x},
\end{align}
where $\eta_i = \log(\lambda_i / \lambda_K)$ is the natural parameter, which conveniently becomes unconstrained real-valued.
Note that, like with $\bold x$ and $\boldsymbol \lambda$, we will drop the $K$th coordinate to denote $\boldsymbol \eta = (\eta_1, \dots, \eta_{K-1})$, since $\eta_K = \log(1) = 0$ is fixed.\footnote{In principle, we could let $\eta_K$ vary together with $\eta_1, \dots, \eta_{K-1}$; Eqs. \ref{eq:norm_const} and \ref{eq:canon} would still hold, since the additional term $e^{\eta_K x_K}$ in the density would compensate the change in $C(\boldsymbol \eta)$. However, such a model would be overparameterized as it would be invariant to a parallel shift across all the $\eta_i$. For mathematical conciseness, we keep $\eta_K$ fixed at 0 and work with $\boldsymbol \eta \in \mathbb R^{K-1}$.}
%, and $\eta_K = 0$.\footnote{Note that the constraint $\sum_i \lambda_i = 1$ has been absorbed into the last component $\eta_K = \log(1) = 0$, which is now fixed, and we denote $\boldsymbol \eta = (\eta_1, \dots, \eta_{K-1})$.}
In this notation, the normalizing constant becomes:
\begin{align} \label{eq:norm_const}
 C(\boldsymbol \eta) =
\sum_{j=1}^K 
\frac
 { e^{ \eta_k }  }
 { \prod_{i \ne k } \left(\eta_k - \eta_i \right)},
\end{align}
which, again, is undefined whenever $\eta_i = \eta_k$ for some $i \ne k$.

\section{Numerical computation of the normalizing constant} \label{sec:numerics}

Eq. \ref{eq:norm_const} can be vectorized efficiently as follows:

\begin{algorithm}
\caption{Vectorized computation of the normalizing constant} \label{alg:vec}
    \hspace*{\algorithmicindent} \textbf{Input:} A parameter vector $\boldsymbol \eta$
    \\
    \hspace*{\algorithmicindent} \textbf{Output:} The normalizing constant $C(\boldsymbol \eta)$    
\begin{algorithmic}[1]
%\For{$i=1, 2, \dots, K-1$}
\State Compute a $K \times K$ matrix of differences $M = [\eta_k - \eta_i]_{i, k=1}^K$ and add to this the identity matrix $I$. A tensor with an additional batch dimension can be used if necessary.
\State Take the product of the rows of $M+I$, using log space as necessary.
\State Multiply the resulting vector componentwise with the vector $[e^{\eta_k}]_{k=1}^K$, and sum up the terms. \label{item:sum}
\end{algorithmic}
\end{algorithm}

%\begin{enumerate}
% \item Compute a $K \times K$ matrix of differences $M = [\eta_k - \eta_i]_{i, k=1}^K$ and add to this the identity matrix $I$. A tensor with an additional batch dimension can be used if necessary.
% \item Take the product of the rows of $M+I$, using log space as necessary.
% \item Multiply the resulting vector componentwise with the vector $[e^{\eta_k}]_{k=1}^K$, and sum up the outputs. \label{item:sum}
%\end{enumerate}

\subsection{Catastrophic cancellation} \label{sec:catastrophic}

Algorithm \ref{alg:vec} is easy to code up %\footnote{See \url{https://github.com/cunningham-lab/cb_and_cc/blob/master/cc/cc_funcs.py}.} 
and adds little computational overhead to most models.
However, the summation in Step \ref{item:sum} involves positive and negative numbers, which can result in catastrophic cancellation \citep{goldberg1991every}. 
\emph{We stress that the log-sum-exp trick, while useful for preventing overflow, {cannot} address catastrophic cancellation (see Section \ref{sec:overflow}).}
For example, consider the case $K=5$ with $\boldsymbol \eta = (1, 2, 3, 4)$. %\footnote{As discussed, $\eta_K = \log(\lambda_K / \lambda_K) = 0$ is fixed. Eq. \ref{eq:norm_const} still holds true for $\eta_K \ne 0$, since the change in $C(\boldsymbol \eta)$ would compensate for the additional term $e^{\eta_K x_K}$ in the density, however this would lead to an overparameterized model.} 
In single-precision floating-point, the summation in Eq. \ref{eq:norm_const} evaluates to:
\begin{align}
\sum_{k=1}^K 
\frac
 { e^{ \eta_k }  }
 { \prod_{i \ne k } \left(\eta_k - \eta_i \right)} \nonumber
 &= 
 -0.45304695 + 1.847264   -3.3475895  + 2.2749228 +  0.04166667
 \\
 &= 0.363217. %15. these last digits are garbage
\end{align}
Note that the output is an order of magnitude smaller than (at least one of) the summands, and as a result we have lost one digit in precision. As we increase the dimension of the CC, the cancellation becomes more severe.
For example, in the case $K=10$ with $\boldsymbol \eta = (1, 2, 3, 4, 5, 6, 7, 8, 9)$, the same summation becomes:
\begin{align} \nonumber
\sum_{k=1}^K 
\frac
 { e^{ \eta_k }  }
 { \prod_{i \ne k } \left(\eta_k - \eta_i \right)}
 &= 
 -6.7417699 \times 10^{-5} -7.3304126\times 10^{-4}  + 4.6494300\times 10^{-3} 
 \\
 & \ \  \ \ \ \ -1.8957691\times 10^{-2}   + 5.1532346\times 10^{-2}  + 9.3386299\times 10^{-2} \nonumber
\\
 & \ \  \ \ \ \ 
 + 1.0879297\times 10^{-1} -7.3932491\times 10^{-2} +  2.2329926\times 10^{-2} \nonumber
  \\
   & \ \  \ \ \ \ 
   -2.7557318\times 10^{-6} \nonumber
\\
&= 3.5982\times 10^{-4}.
\end{align}
We have now lost 3 digits, since the leading summand is 3 orders of magnitude greater than the output.
If we continue increasing $K$ in the same way, by $K=20$ we will have lost 6 digits, and by $K=25$ we are no longer able to compute $C(\boldsymbol \eta)$ to even a single significant digit.
If we were to use double-precision floating-point instead, by $K=40$ we will have lost 13 digits, and by $K=50$ we can no longer compute $C(\boldsymbol \eta)$ to a single significant digit.

We can summarize the problem as follows: \emph{when $C(\boldsymbol \eta)$ is of a much lower order of magnitude than the leading term in the summation of Eq. \ref{eq:norm_const}, numerical computation fails due to catastrophic cancellation}.
To complicate things further, the relationship between the two orders of magnitude (that of $C(\boldsymbol \eta)$ and that of the leading summand) is nontrivial.
This relationship depends on the relative size of the exponential terms $e^{\eta_k}$ and the products of the differences $\eta_k - \eta_i$, and is not straightforward to analyze.
However, if two elements of $\boldsymbol \eta$ are particularly close to one another, meaning that $|\eta_{j_1} - \eta_{j_2}|$ is close to 0 for some $j_1$ and $j_2$, it follows that the corresponding summands: 
\begin{align} \label{eq:j1j2}
\frac
 { e^{ \eta_{j_1} }  }
 { \prod_{i \ne {j_1} } \left(\eta_{j_1} - \eta_i \right)},
 \ \ \mathrm{ and } \ \ 
 \frac
 { e^{ \eta_{j_2} }  }
 { \prod_{i \ne {j_2} } \left(\eta_{j_2} - \eta_i \right)},
\end{align}
are (a) large in magnitude (due to the term $\eta_{j_1} - \eta_{j_2}$ in the denominator), (b) of opposing sign (due to the difference in $\eta_{j_1} - \eta_{j_2}$ versus $ \eta_{j_2} - \eta_{j_1}$), and (c) similar in absolute value (since the terms in the product are approximately equal).
Thus, it is likely that the terms in Eq. \ref{eq:j1j2} are of leading order and catastrophically cancel each other out.
Note that, as the dimensionality $K$ increases, it becomes more likely that \emph{some} pair of the components are close to one another, and therefore computation becomes harder.

\begin{wrapfigure}{r}{0.45\textwidth}
  \begin{center}
  \vspace{-8mm}
    \includegraphics[width=0.57\textwidth]{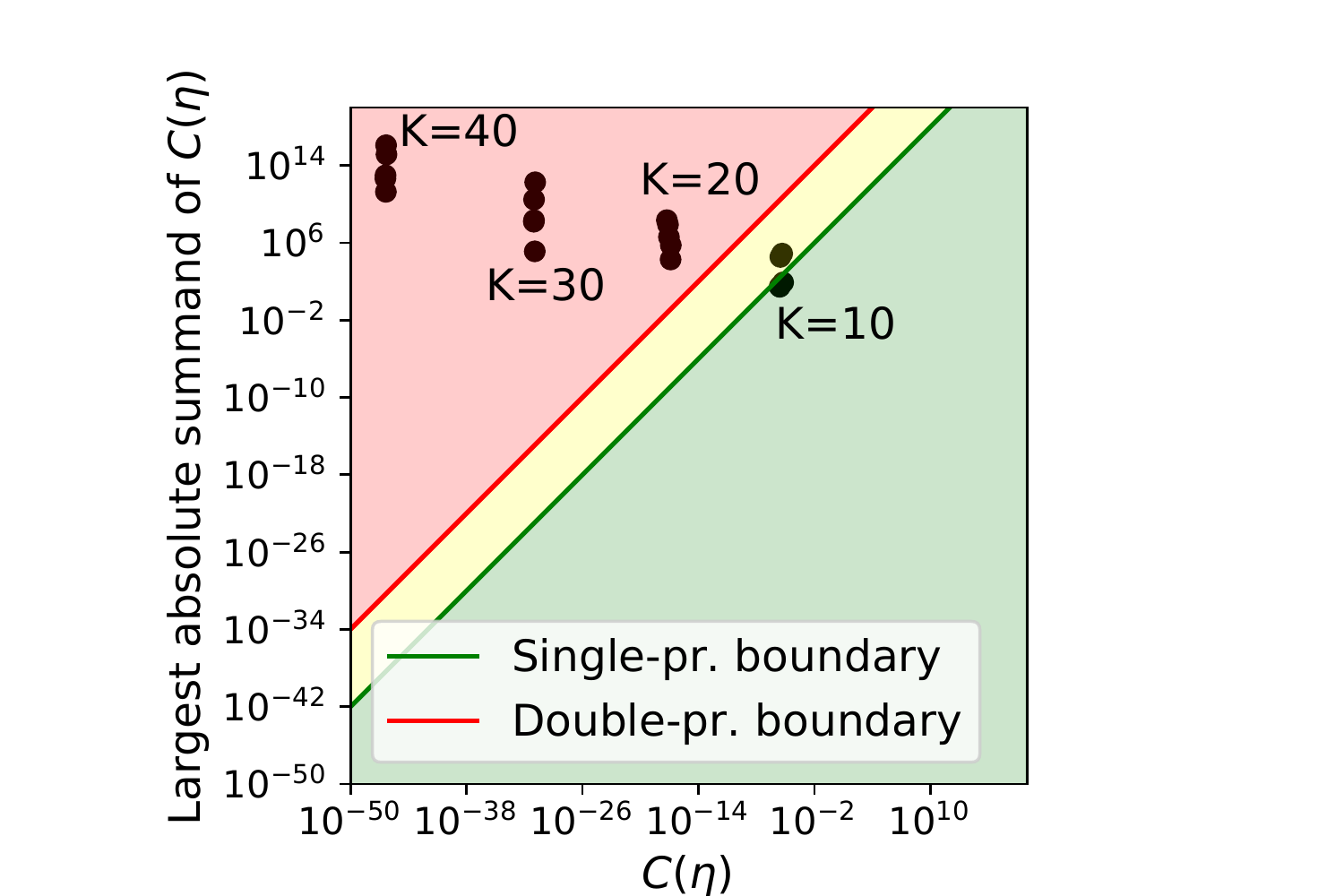} 
  \end{center}
  \caption{Scaling behavior of $C(\boldsymbol \eta)$ relative to its summands (from Eq. \ref{eq:norm_const}), as the dimension $K$ varies. Each point represents a random draw of $\eta_i \overset{iid}{\sim} N(0,1)$, for which we compute $C(\boldsymbol \eta)$. Note that catastrophic cancellation depends on the difference in order of $C(\boldsymbol \eta)$ and its summands; in the green region this difference is at most 8 orders of magnitude, so that single-precision floating-point is sufficient. In the yellow region, it is between 8 and 16 orders of magnitude, so that single-precision fails due to catastrophic cancellation, but double-precision succeeds. In the red region, both fail.
 } \label{fig:0} \vspace{-5mm}
\end{wrapfigure}

Another helpful intuition can be obtained by reasoning from the integral:
\begin{align}
 C(\boldsymbol \eta) =  \int_{\mathbb S^K} e^{\boldsymbol \eta ^\top \bold x} d\mu(\bold x).
\end{align}
As $K$ increases, the Lebesgue measure of the simplex decays like $1/K!$.\footnote{As can be seen, for example, by taking Eq. \ref{eq:dirnorm} with $\alpha_i = 0$ for all $i$.}
Therefore, assuming the components of $\boldsymbol \eta$ are $O(1)$, we have that $e^{\boldsymbol \eta^\top \bold x} = O(1)$ also, and therefore $C(\boldsymbol \eta) = O(1/K!)$.
Under this assumption, we also have that $\eta_k - \eta_i = O(1)$, and therefore the summands in Eq. \ref{eq:norm_const} cannot decay factorially fast (they may, but need not, decay at most exponentially due to the product of $K-1$ terms of constant order in the denominator).
Thus, such a regime implies catastrophic cancellation is inevitable for large enough $K$.
On the other hand, when all the components of $\boldsymbol \eta$ are far from one another, we are spared of such cancellation and Algorithm \ref{alg:vec} succeeds, including in high dimensions.
We demonstrate these behaviors empirically in the following experiments (see Figures \ref{fig:0} and \ref{fig:1}).

\subsubsection{Experiments} \label{sec:experiments}

To evaluate the effectiveness of Algorithm \ref{alg:vec} for computing $C(\boldsymbol \eta)$, we first implemented an oracle capable of correctly computing $C(\boldsymbol \eta)$ to within 4 significant figures (at a potentially large computational cost).
Our oracle is an implementation of Eq. \ref{eq:norm_const} with arbitrary-precision floating-point, using the \texttt{mpmath} library \citep{mpmath}.
In particular, for a given $\boldsymbol \eta$, we ensure the level of precision is set appropriately by computing Eq. \ref{eq:norm_const} repeatedly at increasingly high precision, until the outputs converge (to 4 significant figures).
Equipped with this oracle, we then drew $\boldsymbol \eta$ vectors from a normal prior for a variety of dimensions $K$, to analyze the behavior of $C(\boldsymbol \eta)$.

First, we took $\eta_i \overset{iid}{\sim} N(0,1)$ and compared the magnitude of $C(\boldsymbol \eta)$ to the magnitude of the largest summand in Eq. \ref{eq:norm_const}.
The results are plotted in Figure \ref{fig:0}, where we observe that $C(\boldsymbol \eta)$ decays rapidly in $K$, whereas the same is not true of its (largest) summands.
Thus, under this prior, Algorithm \ref{alg:vec} is unsuccessful except in low-dimensional settings.

\begin{figure}%{r}{0.55\textwidth}
  \begin{center}
    \includegraphics[width=0.65\textwidth]{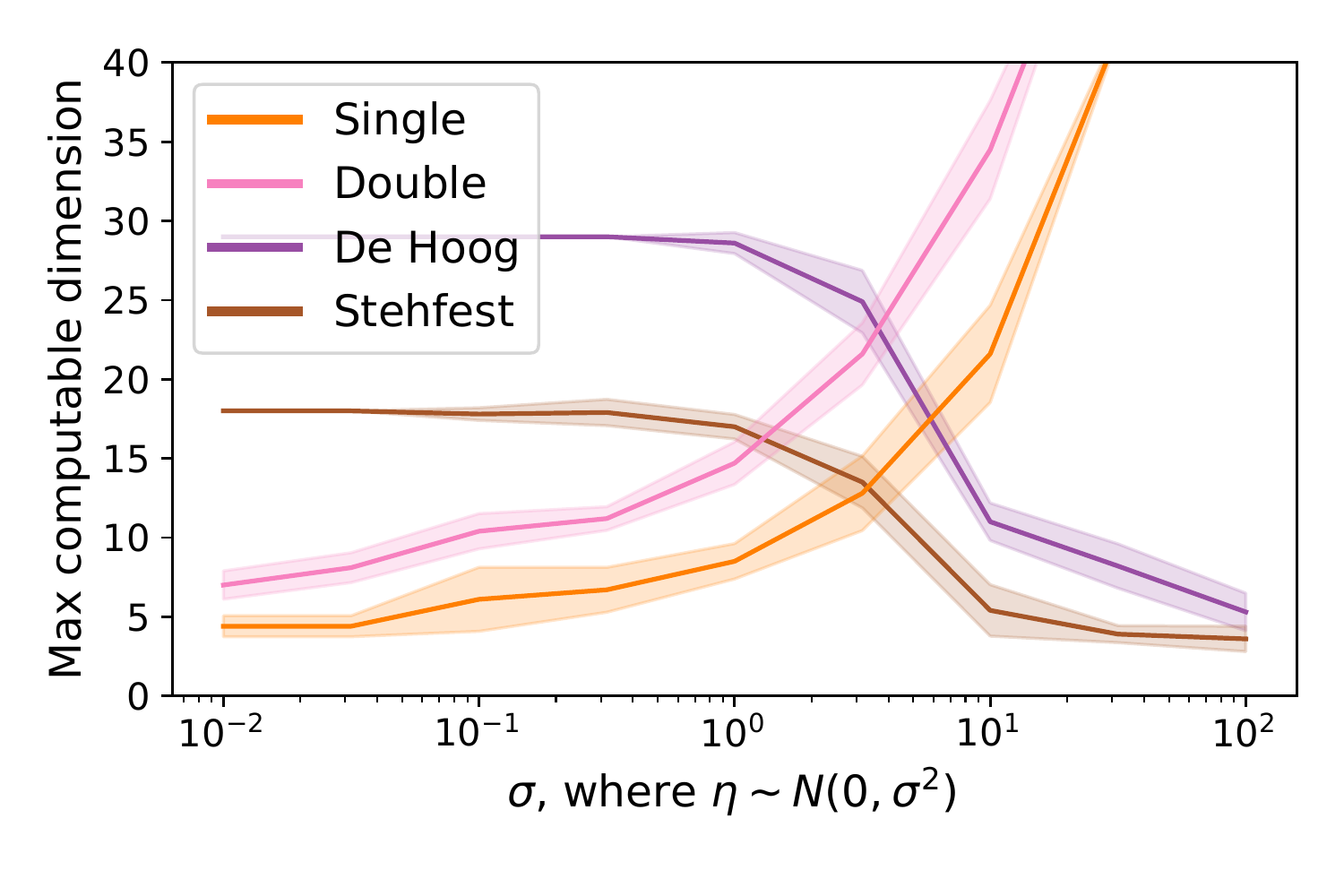} 
  \end{center}
  \caption{Scaling behavior of the numerical accuracy for computing $C(\boldsymbol \eta)$, as a function of the spread of the parameter values. For each $\sigma$ (x-axis), we draw $\eta_i \overset{iid}{\sim} N(0,\sigma^2)$ for $i=1,\dots,40$, and we compute $C(\eta_1, \dots, \eta_{K-1})$, for each $K=3, \dots, 40$. We then show on the y-axis the highest $K$ such that the numerical value of $C(\eta_1, \dots, \eta_{K-1})$ was equal to that obtained by the oracle (to 3 significant figures). The orange and pink lines show the result for Algorithm \ref{alg:vec} using single- and double-precision, respectively. The purple and brown lines show the result for computing $C(\boldsymbol \eta)$ using the inverse Laplace transform, i.e., Eq. \ref{eq:lapinv}, as discussed in Section \ref{sec:laplace}. At each level of $\sigma$, we show error bars over 10 random draws of $\boldsymbol \eta$.
 } \label{fig:1}
\end{figure}

Next, we let the spread of $\boldsymbol \eta$ vary by drawing $\eta_i \overset{iid}{\sim} N(0, \sigma^2)$, where $\sigma$ ranges between $10^{-2}$ and $10^2$.
We then plotted, for each $\sigma$, the highest value of $K$ such that the output of Algorithm \ref{alg:vec} agreed with the Oracle to 3 significant figures.
We repeated the procedure using single- and double-precision floating-point (orange and pink lines in Figure \ref{fig:1}, respectively), as well as two Laplace inversion methods that will be discussed in Section \ref{sec:laplace}.
As $\sigma$ increases, the parameter values tend to move away from one another, making computation easier and allowing for much higher dimensions.
On the other hand, when $\sigma$ decreases, the parameter values come closer, bringing us into the unstable regions of Eq. \ref{eq:norm_const}, and Algorithm \ref{alg:vec} fails for all but just a few dimensions.

\subsection{Inverse laplace transform} \label{sec:laplace}

In this Section, we show that $C(\boldsymbol \eta)$ can be written as the inverse Laplace transform of a function that does not suffer from catastrophic cancellation.
In particular, said function can be passed to Laplace inversion methods \citep{davies1979numerical} in order to evaluate $C(\boldsymbol \eta)$ in the regime where Algorithm \ref{alg:vec} fails.

\textbf{Proposition:} For a function $f:\mathbb R^+ \to \mathbb R^+$, let $\mathcal L[f](s) = \int_0^\infty f(t) e^{-st} dt$ denote the Laplace transform. Define the function $c:\mathbb R^+ \to \mathbb R^+$ by: 
\begin{align} \label{eq:cdef}
c(t) &= \int_{\{\bold x : \sum_{i=1}^{K-1} x_i \le t\}}  \prod_{i=1}^{K} e^{\eta_i x_i} d\mu(\bold x).
\end{align}
Then the Laplace transform of $c$ is equal to:
\begin{align} \label{eq:lap1}
\mathcal{L}[c](s) = \prod_{i=1}^K \frac{1}{ s - \eta_i}.
\end{align}

\textbf{Remark:} The function $c$ includes the normalizing constant of the CC as a special case $C(\boldsymbol \eta) = c(1)$. More generally, we have that:
\begin{align} \label{eq:scaledlap}
c(t) = t^{K-1} C(\boldsymbol \eta / t)^{-1}
,
\end{align}
as can be seen from letting $\tilde x_i = x_i / t$ in the integral \ref{eq:cdef}.

\textbf{Proof:} Let $f_{\eta}(x) = e^{\eta x}$, so that:
\begin{align} \nonumber
c(t) &= \int_{\{\bold x : \sum_{i=1}^{K-1} x_i \le t\}}  \prod_{i=1}^{K} f_{\eta_i}(x_i) d\mu(\bold x)
\\ \label{eq:defnc}
&= \int_0^t \int_0^{t-x_1} \cdots \int_0^{t-x_1-\cdots-x_{K-2}}  \prod_{i=1}^{K} f_{\eta_i}(x_i)
dx_{K-1} \cdots dx_2 dx_1.
\end{align}
Next we apply the transformation from \cite{wolpert1995estimating}, defined as $w_1 = t$, $w_k = w_{k-1} - x_{k-1}$, or equivalently, $w_k = t - \sum_{i=1}^{k-1} x_i$. We can then write our integral as a convolution:
\begin{align} \nonumber
&c(t) = \int_0^{w_1} \int_0^{w_2} \cdots \int_0^{w_{K-1}}  \prod_{i=1}^{K} f_{\eta_i}(x_i)
dx_{K-1} \cdots dx_2 dx_1
\\ \nonumber
&= \int_0^{w_1} \int_0^{w_2}  \hspace{-1mm}  \cdots \int_0^{w_{K-2}}  \prod_{i=1}^{K-2} f_{\eta_i}(x_i) \int_0^{w_{K-1}} f_{\eta_{K-1}}(x_{K-1}) f_{\eta_K}(w_{K-1} - x_{K-1})
dx_{K-1} \cdots dx_2 dx_1
\\ \nonumber
&= \int_0^{w_1} \int_0^{w_2} \cdots \int_0^{w_{K-2}}  \prod_{i=1}^{K-2} f_{\eta_i}(x_i) (f_{\eta_{K-1}} \ast f_{\eta_{K}}) (w_{K-1})
dx_{K-2} \cdots dx_2 dx_1
\\ \nonumber
&= \int_0^{w_1} \hspace{-1mm}  \cdots \int_0^{w_{K-3}}  \prod_{i=1}^{K-3} f_{\eta_i}(x_i) \int_0^{w_{K-2}}  \hspace{-1mm}  f_{\eta_{K-2}}(x_{K-2}) (f_{\eta_{K-1}}  \hspace{-1mm}  \ast f_{\eta_{K}}) (w_{K-2} - x_{K-2})
dx_{K-2} \cdots  dx_1
\\ \nonumber
&= \int_0^{w_1} \int_0^{w_2} \cdots \int_0^{w_{K-3}}  \prod_{i=1}^{K-3} f_{\eta_i}(x_i) (f_{\eta_{K-2}} \ast f_{\eta_{K-1}} \ast f_{\eta_{K}}) (w_{K-2})
dx_{K-3} \cdots dx_2 dx_1
\\ \nonumber
&= \cdots
\\ 
&= ( \circledast _{i=1}^K f_{\eta_i} ) (t)
.
\end{align}
Next, since the Laplace transform of a convolution equals the product of the Laplace transforms, we have that:
\begin{align}
\mathcal{L}[c](s) = \prod_{i=1}^K \mathcal{L}[f_i](s),
\end{align}
but the univariate case is simply: 
\begin{align}
\mathcal{L}[f_i](s) = \int_0^\infty e^{(\eta_i - s)t} dt = \frac{1}{s - \eta_i},
\end{align}
and the result follows. \qed

\textbf{Corollary:} The normalizing constant of the continuous categorical distribution can be written as the following inverse Laplace transform:
\begin{align} \label{eq:lapinv}
C(\boldsymbol \eta) = \mathcal{L}^{-1}\left[ \prod_{i=1}^K \frac{1}{s - \eta_i} \right] (1).
\end{align}

\textbf{Proof:} Take the inverse Laplace transform in Eq. \ref{eq:lap1} to find:
\begin{align} 
c(t) = \mathcal{L}^{-1}\left[ \prod_{i=1}^K \frac{1}{s - \eta_i} \right] (t).
\end{align}
Taking $t=1$ gives the desired result. \qed

Unlike Eq. \ref{eq:norm_const}, the product in Eq. \ref{eq:lap1} does not suffer from catastrophic cancellation, nor does it diverge whenever $\eta_{j_1} \approx \eta_{j_2}$ for some $j_1 \ne j_2$.
The corresponding Laplace inversion, i.e., Eq. \ref{eq:lapinv}, provides an alternative method to compute our normalizing constant $C(\boldsymbol \eta)$.
Numerous numerical algorithms exist for inverting the Laplace transform; see \citep{cohen2007numerical} for a survey. 
We note, however, that inverting the Laplace transform is generally a hard problem \citep{epstein2008bad}.

%\textbf{Experiment:} 
Empirically, we found some modest empiricasuccess in computing Eq. \ref{eq:lapinv} numerically.
We tested three inversion algorithms, due to Talbot \citep{talbot1979accurate}, Stehfest \citep{stehfest1970algorithm}, and De Hoog \citep{de1982improved}.
The experimental setup was identical to that of Section \ref{sec:experiments}, and the results are incorporated into Figure \ref{fig:1}.
We found De Hoog's method to be the most effective on our problem, whereas Talbot's always failed and is omitted from the Figure.
Importantly, De Hoog's method showed some success in the regime where Algorithm \ref{alg:vec} failed, meaning it could be used as a complementary technique.
However, no inversion method was able to scale beyond $K=30$.

\subsection{Inductive approach} \label{sec:inductive}

In this section, we provide an alternative algorithm to compute our normalizing constant. 
This algorithm will be based on the following recursive property of $C(\boldsymbol \eta)$, which was implicitly used as part of the proof of Eq. \ref{eq:norm_const}  \citep{gordon2020continuous}.

\textbf{Proposition:} Define the subvector notation $\boldsymbol \eta_{:k} = (\eta_1, \dots, \eta_{k-1})$, and make the dependence on $K$ explicit by writing $C_K(\boldsymbol \eta) = C(\eta_1, \dots, \eta_{K-1})$. We also use the notation $\boldsymbol \eta - \eta_k = (\eta_1 - \eta_k, \dots, \eta_{K-1} - \eta_k)$. Then:
\begin{align} \label{eq:recursion}
C_K(\boldsymbol \eta) = \frac{e^{\eta_{K-1}} C_{K-1}\left(\boldsymbol \eta_{:(K-1)} - \eta_{K-1}\right) - C_{K-1}\left(\boldsymbol \eta_{:(K-1)}\right)}{\eta_{K-1}}
.
\end{align}

\textbf{Proof:} We start from the integral definition of the left hand side:
\begin{align} \nonumber
C_K(\text{\boldmath$\eta$})
 &= \int_{{\mathbb{S}^{K-1}}} e^{\boldsymbol \eta^\top \bold x} d\mu
\\
&= \int_0^1 \int_0^{1-x_1} \cdots \int_0^{1-x_1-\cdots-x_{K-2}} e^{\sum_{i=1}^{K-1} \eta_i x_i }
dx_{K-1} \cdots dx_2 dx_1.
%\\
%&= \int_0^1 \int_0^{1-x_1} \cdots \int_0^{1-x_1-\cdots-x_{K-2}} e^{\sum_{i=1}^{K-2} \eta_i x_i } e^{\eta_{K-1} x_{K-1}} dx_{K-1} \cdots dx_2 dx_1.
\end{align}
For the innermost integral, we have:
\begin{align} \nonumber
\int_0^{1-x_1-\cdots-x_{K-2}} e^{\sum_{i=1}^{K-1} \eta_i x_i }
dx_{K-1} 
&=
e^{\sum_{i=1}^{K-2} \eta_i x_i} \int_0^{1-x_1-\cdots-x_{K-2}} 
e^{\eta_{K-1} x_{K-1}} dx_{K-1} 
\\ \nonumber
&= 
e^{\sum_{i=1}^{K-2} \eta_i x_i} 
\left( \frac{ e^{\eta_{K-1}(1-x_1-\cdots-x_{K-2})} - 1 }{\eta_{K-1}} \right) 
\\ \nonumber
&= 
 \frac{ e^{\eta_{K-1}} e^{\sum_{i=1}^{K-2} ( \eta_i - \eta_{K-1}) x_i}  - e^{\sum_{i=1}^{K-2} \eta_i x_i}  }{\eta_{K-1}} 
\\
 &= \label{eq:inner}
 \frac{ e^{\eta_{K-1}} e^{ ( \boldsymbol \eta_{:(K-1)} - \eta_{K-1})^\top \bold x_{:(K-1)}}  - e^{\boldsymbol \eta_{:(K-1)} ^\top \bold x_{:(K-1)} }  }{\eta_{K-1}} 
,
\end{align}
and the result follows by linearity of the integral. \qed

\textbf{Remark:} The base case $K=2$ corresponds to the univariate continuous Bernoulli distribution, which admits a straightforward Taylor expansion that is useful around the unstable region $\eta \approx 0$ \citep{loaiza2019continuous}:
\begin{align} 
C_2(\boldsymbol \eta_{:2}) = \frac{e^{\eta_1} - 1}{\eta_1} =  \frac{(1 + \eta_1 + \frac{1}{2!} \eta_1^2 + \cdots) - 1}{\eta_1} = 1 + \frac{1}{2!} \eta_1 + \cdots
.
\end{align}

In words, Eq. \ref{eq:recursion} is stating that we can compute $C(\boldsymbol \eta)$ for the full parameter vector $\boldsymbol \eta = (\eta_1, \dots, \eta_{K-1})$ by first computing it for the lower-dimensional parameter vectors:
\begin{align*}
\boldsymbol \eta_{:K-1} &= (\eta_1, \dots, \eta_{K-2}),
\\
\boldsymbol \eta_{:K-1} - \eta_{K-1} &= (\eta_1 - \eta_{K-1}, \dots, \eta_{K-2} - \eta_{K-1})
.
\end{align*}
Substituting these back into Eq. \ref{eq:recursion}, we have that:
\begin{align*}
C_{K-1}(\boldsymbol \eta_{:K-1}) &= \frac{e^{\eta_{K-2}} C_{K-2}(\boldsymbol \eta_{1:K-2} - \eta_{K-2})  - C_{K-2}(\boldsymbol \eta_{1:K-2})}{\eta_{K-2}},
\\
C_{K-1}(\boldsymbol \eta_{:K-1} - \eta_{K-1}) &= \frac{e^{\eta_{K-2}-\eta_{K-1}} C_{K-2}(\boldsymbol \eta_{1:K-2} - \eta_{K-2})  - C_{K-2}(\boldsymbol \eta_{1:K-2} - \eta_{K-1}) }{\eta_{K-2}-\eta_{K-1}}
.
\end{align*}
Note that we are now left with not 4, but 3 new parameter vectors to recurse on: $\boldsymbol \eta_{1:K-3}$, $\boldsymbol \eta_{1:K-3} - \eta_{K-1}$, and $\boldsymbol \eta_{1:K-3} - \eta_{K-2}$.
Repeating the argument $K-2$ times and working backwards we obtain Algorithm \ref{alg:ind} for computing the normalizing constant.
%More generally, equation \ref{eq:recursion} is saying that, from each parameter vector we feed into the LHS we obtain two new parameter vectors, one where we simply `lob off' the last element, and another where we subtract the last component from the entire vector.
%The latter operation results in the same new subvector from all current parameter vectors to which it is applied.
%Thus, at each step we only increase the parameter count by 1.
%After applying $K-2$ steps, we are left with $K-1$ parameter vectors of length 1 (of the form $\eta_1 - \eta_k$, for $k=2,\dots,K$), to which we can apply the base case \ref{eq:basecase}.
%These can be fed back through equation \ref{eq:recursion} over $K-2$ steps to obtain $C_K(\boldsymbol \eta)^{-1}$, the first step requiring $K-2$ evaluations, then $K-3$, then $K-4$ and so forth.
\begin{algorithm}
\caption{Inductive computation of the normalizing constant} \label{alg:ind}
    \hspace*{\algorithmicindent} \textbf{Input:} A parameter vector $\boldsymbol \eta$
    \\
    \hspace*{\algorithmicindent} \textbf{Output:} The normalizing constant $C(\boldsymbol \eta)$    
\begin{algorithmic}[1]
%\For{$i=1, 2, \dots, K-1$}
\State Initialize $\bold c = (1, \dots, 1) \in \mathbb R^{K-1}$ and set $\bold{\tilde c} = \bold c$.
%\EndFor
\For{$k=1, 2, \dots, K-1$}
\For{$i=1,\dots,K-k$}
\State Set $\xi_i = \eta_k - \eta_{k+i}$
\State Set $\tilde c_i = \frac{e^{\xi_i} c_1 - c_{i+1}}{\xi_i}$
\EndFor
%\For{$i=1,\dots,K-k$}
\State Set $\bold c = \bold{ \tilde c}$ %$C_i = \tilde C_i$ for all $i$
%\EndFor
\EndFor \\
\Return $c_1$
\end{algorithmic}
\end{algorithm}

We find the numerical properties of Algorithm \ref{alg:ind} to perform similarly to Algorithm \ref{alg:vec}, suffering from the same cancellation issues in high dimensions. 
Nevertheless, we hope this alternative scheme may help to inspire further numerical improvements.
For example, since the floating-point behavior of Algorithm \ref{alg:ind} depends on the ordering of the elements of $\boldsymbol \eta$, it may be possible do design a reordering scheme that improves the overall precision of the algorithm; such reorderings have been explored in the context of sampling algorithms for the CC \citep{gordon2020continuous}.
Other possibilities include Kahan summation \citep{kahan1965pracniques} or the compensated Horner algorithm \citep{langlois2007ensure}; we leave their study to future work.

%\section{Interval arithmetic}
%
%One way to diagnose the level of error in our normalizing constant is through interval arithmetic (IEEE standard). 
%Note there are successful examples of using this in chained computation (e.g. NN), (Oala et al 2020, Interval Neural Networks).

\subsection{Overflow} \label{sec:overflow}

We conclude this section with a small remark on numerical overflow in the context of Algorithm \ref{alg:vec}.
In high dimensions and with high $\eta$ values, overflow can occur for the terms in the summand, due to a large $e^{\eta_j}$ term or a small denominator.
This can be addressed by re-writing the normalizing constant in a form that allows us to take advantage of the log-sum-exp trick:
\begin{align*} 
\log C(\text{\boldmath$\eta$}) &=
\log \left(
 \sum_{k=1}^K 
\frac
 { { e^{\eta_k}} }
 { \prod_{i \ne k } \left(\eta_k - \eta_i\right)}
 \right)
 \\
 &= 
 \log \left(
 \sum_{k=1}^K 
\frac
 { { e^{\eta_k}} }
 {\mathrm{sign}\left( \prod_{i \ne k } \left(\eta_k - \eta_i\right) \right)  \prod_{i \ne k } |\eta_k - \eta_i| }
 \right)
 \\
 &=
 \log \left(
 \sum_{k=1}^K 
 \mathrm{sign}\left( \prod_{i \ne k } \left(\eta_k - \eta_i\right) \right) 
  \exp \left( \eta_k -  \sum_{i \ne k } \log \left|\eta_k - \eta_i \right| \right)
  \right)
  .
\end{align*}

%In this formulation, each of the $K$ exponents, $ \left( \eta_k -  \sum_{i \ne k } \log \left|\eta_i - \eta_k\right| \right)$, are typically of a similar order of magnitude, but taking the exponential sends them to very different orders.
%In the flat-distribution setting, higher floating point precision is required for the exponents as well as the resulting summands in order to faithfully recover the normalizing constant.

\section{Normalizing constant with repeated parameters} \label{sec:repeat}

Whenever we have an equality between any pair of parameters, Eq. \ref{eq:norm_const} is undefined and, indeed, its proof by \cite{gordon2020continuous} breaks down.
In this Section, we derive a counterpart to Eq. \ref{eq:norm_const} for the case when 2 or more elements of $\boldsymbol \eta$ are equal to one another.
Note that, for mathematical convenience, we shall now denote $\boldsymbol \eta \in \mathbb R^K$, where the $K$th component $\eta_K$ is now included in the vector $\boldsymbol \eta$.
As discussed in Section \ref{sec:intro}, this component can be taken as fixed at 0, or it can be treated as an additional free parameter, resulting in an overparameterized model (our results will remain correct nevertheless).

\subsection{A simple example} \label{sec:simex}

First, we illustrate the main idea of our argument using an example with $K=3$ and $\eta_1 = \eta_2 \ne \eta_3 = 0$.\footnote{Note that the case $\eta_3 \ne 0$ and the case $\eta_1 \ne \eta_2 = \eta_3$ are mathematically equivalent (albeit more algebraically cumbersome), since we can permute the elements of $\boldsymbol \eta$ and shift by a constant without loss of generality.}
By definition, the normalizing constant is then:
\begin{align}
C_3(\boldsymbol \eta) = \int_0^1 \int_0^{1-x_1} e^{\eta_1 x_1 + \eta_2 x_2 } dx_2 dx_1 = \int_0^1 \int_0^{1-x_1} e^{\eta_1 ( x_1 + x_2) } dx_2 dx_1 .
\end{align}
We  apply the change of variables $u=x_1 + x_2$, $v = x_1$ to obtain:
\begin{align}
C_3(\boldsymbol \eta) = \int_0^1 \int_0^{u} e^{\eta_1 u } dv du =  \int_0^1 u e^{\eta_1 u} du %= C_2(\eta_1) \mathbb E_{u \sim \mathcal{CC}(\eta_1)}[u]
%= C_2(\eta_1) \int_0^1 u  \cdot \frac{e^{\eta_1 u}}{C_2(\eta_1)} du
.
\end{align}
Note that the last integral corresponds to the expectation of a univariate CC random variable, an idea we now generalize.
%, we can compute the normalizing constant by computing expectations with respect to a CC distribution of lower dimension.

\subsection{General formula} \label{sec:general}

We start by proving the following lemma, which relates $C(\boldsymbol \eta)$, where some elements of $\boldsymbol \eta$ are repeated (potentially many times), to $C(\boldsymbol \eta')$, where $\boldsymbol \eta'$ collapses the repeated elements of $\boldsymbol \eta$ onto a single coordinate.
% says that if we have a $\mathcal{CC}(\boldsymbol \eta)$ where some elements of $\boldsymbol \eta$ are repeated (potentially many times), then we can write expectations of functions in terms of lower-dimensional CC terms, provided the functions do not depend on coordinates of $\boldsymbol x$ corresponding to repeated components of $\boldsymbol \eta$.
We again use subvector notation $\bold x_{k:} = (x_k, x_{k+1}, \dots, x_{K-1})$.

\textbf{Lemma}: Let $\bold x \sim \mathcal{CC}_K(\boldsymbol \eta)$ with  $\eta_1=\eta_2=\dots=\eta_{k}$, for $1 \leq k \leq K-1$, and let $f:\mathbb R^{K-k-1} \to \mathbb R$. Then:
\begin{equation}
{C_K(\boldsymbol \eta)} \mathbb{E}_{\bold x \sim \mathcal{CC}_K(\boldsymbol \eta)}[f(\bold x_{(k+1):})] = {C_{K-k+1}(\boldsymbol \eta_{k:})} \mathbb{E}_{\bold u \sim \mathcal{CC}_{K-k+1}(\boldsymbol\eta_{k:})}\left[\dfrac{u_1^{k-1}}{(k-1)!} f(\bold u_{2:})\right].
\end{equation}
%where $v_{j:}$ denotes the vector whose entries are the coordinates of $v$, from coordinate $j+1$ onward.

\textbf{Remark:} We are not assuming that the last $K-k$ coordinates of $\boldsymbol\eta$ are all different, we are simply assuming that the first $k$ are identical.
Note also that the positions of the coordinates could be arbitrary and need not be the first $k$ ones; we can always use this lemma to collapse repeated parameter values provided that the function $f$ does not depend on the corresponding coordinates (we can simply relabel the coordinates by a suitable permutation without loss of generality).
%if the repeated coordinates do not correspond to the first $k$ ones, we can permute these coordinates to the first $k$ positions without loss of generality.
%Before proving this result, some important notes: first, we are not assuming that the last $K-k-1$ coordinates of $\eta$ are all different or that they are different than the first $k$, we are simply assuming that the first $k$ are identical. Second, we are obviously interested in the case where $k$ coordinates are identical, but without loss of generality we can assume it is the first $k$ ones.

\textbf{Proof}: By definition of the left hand side:
\begin{align} \nonumber
C_K&(\boldsymbol \eta) \mathbb{E}_{\bold x \sim \mathcal{CC}_K(\boldsymbol \eta)}[f(\bold x_{(k+1):})] 
=
 \int_0^1 \hspace{-1mm} \int_0^{1-x_1} \hspace{-1mm} \cdots  \int_0^{1-x_1-\cdots -x_{K-2}}  \hspace{-1mm} f(\bold x_{(k+1):}) e^{\boldsymbol \eta ^\top \bold x } dx_{K-1} \cdots dx_2 dx_1 %_1(x_1+x_2+\cdots+x_k) + \eta_{k+1} x_{k+1} + \eta_{k+2}x_{k+2}+ \cdots + \eta_K x_K } 
 \\
 &= \int_0^1 \int_0^{1-x_1}\cdots \int_0^{1-x_1-\cdots -x_{K-2}} f(\bold x_{(k+1):}) e^{\eta_k(x_1 + \dots + x_{k}) + \boldsymbol \eta_{(k+1):} ^\top \bold x_{(k+1):} } dx_{K-1} \cdots dx_2 dx_1. %_1(x_1+x_2+\cdots+x_k) + \eta_{k+1} x_{k+1} + \eta_{k+2}x_{k+2}+ \cdots + \eta_K x_K } 
\end{align}
Consider the following change of variable (note this is just like Section \ref{sec:simex}, but with more bookkeeping):
\begin{equation}
\begin{cases}
u_1 = x_1+x_2+\cdots+x_{k}\\
u_2 = x_{k+1}\\
u_3 = x_{k+2}\\
\hspace{16pt}\vdots \\
u_{K-k} = x_{K-1} \\
v_1 = x_1 \\
v_2= x_2 \\
\hspace{16pt}\vdots \\
v_{k-1} = x_{k-1}
\end{cases}
\end{equation}
This change of variable amounts to an invertible linear transformation with the property that the absolute value of the determinant of its Jacobian is $1$, so that we have:
\begin{align} \label{ugly_int1}
C_K(\boldsymbol \eta) &\mathbb{E}_{\bold x \sim \mathcal{CC}_K(\boldsymbol \eta)}[f(\bold x_{(k+1):})] 
 = \int_0^1 \int_0^{1-u_1} \cdots \int_0^{1-u_1-u_2-\cdots -u_{K-k}} \int_0^{u_1} \int_0^{u_1-v_1} \cdots
 \\
 & \dots \int_0^{u_1 - v_1 - v_2 - \dots - v_{k-2}}  f(\bold u_{2:}) e^{\eta_k u_1 + \boldsymbol \eta_{(k+1):}^\top \bold u_{2:}} dv_{k-1} \cdots dv_2 dv_1 du_{K-k} \cdots du_2 du_1 . \nonumber
\end{align}
Note that the change of variables is such that the integrand does not depend on $v_{1}, v_2, \dots, v_{k-1}$. Therefore:
\begin{align}\label{ugly_int2}
C_K(\boldsymbol \eta) &\mathbb{E}_{\bold x \sim \mathcal{CC}_K(\boldsymbol \eta)}[f(\bold x_{(k+1):})] 
\\
& = \int_0^1 \int_0^{1-u_1} \cdots \int_0^{1-u_1-u_2-\cdots -u_{K-k}} g(u_1) f(\bold u_{2:})  e^{\boldsymbol \eta_{k:}^\top \bold u}  du_{K-k} \cdots du_2 du_1 ,  \nonumber
%& C_K^{-1}(\eta) \mathbb{E}_{x \sim \mathcal{CC}_K(\eta)}[f(x_{k:})]\\
%& =\displaystyle \int_0^1 \int_0^{1-u_1} \int_0^{1-u_1-u_2} \cdots \int_0^{1-u_1-u_2-\cdots -u_{K-k}} f(u_{1:}) \eta_1^{u_1} \eta_{k+1}^{u_2} \eta_{k+2}^{u_3} \dots \eta_K^{u_{K-k+1}} \nonumber\\
%& \int_0^{u_1} \int_0^{u_1-v_1} \cdots \int_0^{u_1 - v_1 - v_2 - \cdots - v_{k-2}}dv_{k-1} \cdots dv_2 dv_1 du_{K-k+1} \cdots du_3 du_2 du_1 \nonumber \\
%& = \int_0^1 \int_0^{1-u_1} \int_0^{1-u_1-u_2} \cdots \int_0^{1-u_1-u_2-\cdots -u_{K-k}} I_{k-1}(u_1)f(u_{1:}) \eta_1^{u_1} \eta_{k+1}^{u_2} \eta_{k+2}^{u_3} \dots \eta_K^{u_{K-k+1}} du_{K-k+1} \cdots du_3 du_2 du_1 \nonumber \\
%& = \mathbb{E}_{u \sim \mathcal{CC}(\eta_1, \eta_{k+1}, \eta_{k+2}, \dots, \eta_K)}[I_{k-1}(u_1) f(u_{1:})]C_{K-k+1}^{-1}(\eta_1, \eta_{k+1}, \eta_{k+2}, \dots, \eta_K) \nonumber
\end{align}
where:
\begin{align}
g(u_1) =  \int_0^{u_1} \int_0^{u_1-v_1} \cdots \int_0^{u_1 - v_1 - v_2 - \cdots - v_{k-2}}dv_{k-1} \cdots dv_2 dv_1.
\end{align}
But this is simply the Lebesgue measure of a simplex inscribed in the hypercube $[0, u_1]^{K-1}$, so that $g(u_1) = u_1^{k-1} \mu(\mathbb S^K) = u_1^{k-1} / (k-1)!$ (this can also be seen by changing variables to $\tilde v_i = v_i / u_i$, or by applying Eq. \ref{eq:scaledlap}). 
Multiplying and dividing by $C_{K-k+1}(\boldsymbol \eta_{k:})$ gives the desired result. \qed

We can now derive a formula for the normalizing constant for an arbitrary parameter vector $\boldsymbol \eta$.

\textbf{Corollary}: Let $\boldsymbol \eta \in \mathbb{R}^{K}$ contain $D \le K$ unique elements. Assume, without loss of generality, that $\boldsymbol \eta = (\eta_1, \dots, \eta_1, \eta_2, \dots, \eta_2, \dots, \eta_D, \dots, \eta_D)$, where each coordinate is repeated $1 \leq r_i \le K$ times, with $\sum_{i=1}^D r_i = K$. Then:
\begin{equation}\label{norm_const_gen}
C_K(\boldsymbol \eta) = {C_D(\eta_1, \eta_2, \dots, \eta_D)}{\mathbb{E}_{\bold u \sim \mathcal{CC}_D(\eta_1, \eta_2, \dots, \eta_D)}\left[\displaystyle \prod_{i=1}^D \dfrac{u_i^{r_i-1}}{(r_i - 1)!}\right]}
\end{equation}

%\textbf{Remark:} The term $C_D(\eta_1, \dots, \eta_D)$ can be evaluated with Eq. \ref{eq:norm_const}, since all the parameter values are distinct.
%The expectation term can be computed by differentiating the moment generating function of $\bold u \sim \mathcal{CC}_D(\eta_1, \eta_2, \dots, \eta_D)$, as discussed by \cite{gordon2020continuous}.
%However, in real data applications exact equality between parameter values may never occur, and it is unclear how close the elements of $\boldsymbol \eta$ should be to be considered equal, if at all.
%Nevertheless, Eq. \ref{norm_const_gen} is important for theoretical completeness.
%Before proving this result, we make two important notes. The first one is that the normalizing constant on the right hand side is with respect to a continuous categorical whose parameter contains only different coordinates, so that the normalizing constant formula from the paper can be used. The second is that the expectation can be computed by differentiating the moment generating function of a CC whose parameter does not have any repeated values. That is: we can compute the right hand side of equation \ref{norm_const_gen}.

\textbf{Proof}:
The result follows by applying the lemma $D$ times. First, we apply the lemma with $f(\cdot) = 1$:
\begin{align}
C_K(\boldsymbol \eta) = {C_{K-r_1+1}( \boldsymbol \eta')}{\mathbb{E}_{\bold u \sim \mathcal{CC}_{K-r_1+1}(\boldsymbol\eta')}\left[\dfrac{u_1^{r_1-1}}{(r_1-1)!}\right]},
\end{align}
where $\boldsymbol \eta' = \boldsymbol \eta_{r_1:} = (\eta_1, \eta_2, \dots, \eta_2, \dots, \eta_D, \dots, \eta_D)$, i.e., we have collapsed the first parameter value onto a single coordinate.
Next, we apply the lemma a second time on the new expectation term to collapse the $\eta_2$ values, this time using $f(u_1)=u_1^{r_1-1}/(r_1-1)!$, which does not depend on the $\eta_2$ coordinates:
\begin{align}
{C_{K-r_1+1}( \boldsymbol \eta')} \mathbb{E}&_{\bold u \sim \mathcal{CC}_{K-r_1+1}(\boldsymbol \eta')} \left[\dfrac{u_1^{r_1-1}}{(r_1-1)!}\right] 
\\ \nonumber
&=
{C_{K-r_1-r_2+2}(\boldsymbol \eta'')}
\mathbb{E}_{\bold u \sim \mathcal{CC}_{K-r_1 - r_2 +2}(\boldsymbol \eta'')}\left[\dfrac{u_1^{r_1-1}}{(r_1-1)!}\dfrac{u_2^{r_2-1}}{(r_2-1)!}\right],
\end{align}
where $\boldsymbol \eta'' = (\eta_1, \eta_2, \eta_3, \dots, \eta_3, \dots, \eta_D, \dots, \eta_D)$.
Repeating $D$ times yields the desired result.
% applying the lemma $D$ times yields:
%\begin{align}
%C_K(\boldsymbol \eta) & = \dfrac{C_{K-r_1-r_2+2}(\eta_1, \eta_2, \eta_{r_1+r_2:})}{\mathbb{E}_{u \sim \mathcal{CC}_{K-r_1-r_2+2}(\eta_1, \eta_2, \eta_{r_1+r_2:})}\left[\dfrac{u_1^{r_1-1}}{(r_1-1)!}\dfrac{u_2^{r_2-1}}{(r_2-1)!}\right]} \\
%& = \dots = \dfrac{C_D(\eta_1, \eta_2, \dots, \eta_D)}{\mathbb{E}_{u \sim \mathcal{CC}_D(\eta_1, \eta_2, \dots, \eta_D)}\left[\displaystyle \prod_{i=1}^D \dfrac{u_i^{r_i-1}}{(r_i - 1)!}\right]} \nonumber
%\end{align}
%as required. 
\qed

Note that Eq. \ref{norm_const_gen} can be computed using known results.
The term $C_D(\eta_1, \dots, \eta_D)$ can be evaluated with Eq. \ref{eq:norm_const}, since all the parameter values are distinct.
The expectation term can be computed by differentiating the moment generating function of $\bold u \sim \mathcal{CC}_D(\eta_1, \eta_2, \dots, \eta_D)$, as discussed by \cite{gordon2020continuous}.
Note that in real data applications exact equality between parameter values may never occur, and it is unclear how close the elements of $\boldsymbol \eta$ should be in order to warrant applying Eq. \ref{norm_const_gen}.
Nevertheless, Eq. \ref{norm_const_gen} is important for theoretical completeness.

\section{Discussion}

The normalizing constant of the CC distribution is essential to applications, being a necessary prerequisite for evaluating likelihoods, optimizing models, and simulating samples alike.
Computing this normalizing constant is nontrivial, and doing so in high dimensions remains an open problem.
Our work represents a significant step toward this goal, improving our understanding of the numerical properties of different computation techniques, as well as advancing the underlying theory and algorithms.
In addition, we hope our results will help to inspire further advances and to develop increasingly robust numerical techniques that will ultimately enable the use of the CC distribution with arbitrary parameter values on high-dimensional problems.

\bibliography{references}
\bibliographystyle{plainnat}

%\appendix

%\section{Appendix}
%
%
%Optionally include extra information (complete proofs, additional experiments and plots) in the appendix.
%This section will often be part of the supplemental material.

\end{document}